\documentclass[11pt,twoside]{article}
\usepackage[latin9]{inputenc}
\usepackage{array}
\usepackage{multirow}
\usepackage{graphicx}

\makeatletter

\providecommand{\tabularnewline}{\\}

\usepackage{jmlr2e}
\usepackage{amsbsy}
\usepackage{microtype}

\renewcommand{\cite}{\citep}

\ShortHeadings{Preterm Birth Prediction}{Tran, PhD and Luo, MD and Phung, PhD and Morris, MD and Rickard and Venkatesh, PhD}
\firstpageno{1}

\@ifundefined{showcaptionsetup}{}{%
 \PassOptionsToPackage{caption=false}{subfig}}
\usepackage{subfig}
\makeatother

\begin{document}

\title{Preterm Birth Prediction: Deriving Stable and Interpretable Rules
from High Dimensional Data\thanks{This work is partially supported by the Telstra-Deakin Centre of Excellence
in Big Data and Machine Learning.}}

\author{\name{T}ruyen Tran \email truyen.tran@deakin.edu.au \\
 \addr Center of Pattern Recognition and Data Analytics\\
Deakin University, Geelong, VIC 3216, Australia\vspace{-2mm}
\\
\AND\name{W}ei Luo \email wei.luo@deakin.edu.au \\
\addr Center of Pattern Recognition and Data Analytics\\
Deakin University, Geelong, VIC 3216, Australia\\
\AND\name{D}inh Phung \email dinh.phung@deakin.edu.au \\
\addr Center of Pattern Recognition and Data Analytics\\
Deakin University, Geelong, VIC 3216, Australia\\
\AND \name{J}onathan Morris \email jonathan.morris@sydney.edu.au\\
 \addr Sydney Medical School, The University of Sydney\\
St Leonards, NSW 2065, Australia\\
\AND \name{K}risten Rickard \email kristen.rickard@sydney.edu.au\\
\addr Clinical and Population Perinatal Health Research\\
Royal North Shore Hospital, St Leonards, NSW 2065 , Australia\\
\AND \name{S}vetha Venkatesh \email svetha.venkatesh@deakin.edu.au\\
\addr Center of Pattern Recognition and Data Analytics\\
Deakin University, Geelong, VIC 3216, Australia}
\maketitle
\begin{abstract}
 Preterm births occur at an alarming rate of 10-15\%. Preemies have
a higher risk of infant mortality, developmental retardation and long-term
disabilities. Predicting preterm birth is difficult, even for the
most experienced clinicians. The most well-designed clinical study
thus far reaches a modest sensitivity of 18.2\textendash 24.2\% at
specificity of 28.6\textendash 33.3\%. We take a different approach
by exploiting databases of normal hospital operations. We aims are
twofold: (i) to derive an easy-to-use, interpretable prediction rule
with quantified uncertainties, and (ii) to construct accurate classifiers
for preterm birth prediction. Our approach is to automatically generate
and select from hundreds (if not thousands) of possible predictors
using stability-aware techniques. Derived from a large database of
15,814 women, our simplified prediction rule with only 10 items has
sensitivity of 62.3\% at specificity of 81.5\%.
\end{abstract}
\global\long\def\bv{\mathbf{v}}
 \global\long\def\bh{\mathbf{h}}
 \global\long\def\ba{\mathbf{a}}
 \global\long\def\ab{\mathbf{a}}
 \global\long\def\bb{\mathbf{b}}
 \global\long\def\Sb{\mathbf{S}}
 \global\long\def\bS{\mathbf{S}}
 \global\long\def\bw{\mathbf{w}}
 \global\long\def\wb{\mathbf{w}}

\global\long\def\vb{\mathbf{v}}
 \global\long\def\hb{\mathbf{h}}
 \global\long\def\Real{\mathbb{R}}
 \global\long\def\Data{\mathcal{D}}
 \global\long\def\Normal{\mathcal{N}}
 \global\long\def\Loss{\mathcal{L}}
 \global\long\def\thetab{\boldsymbol{\theta}}

\global\long\def\LL{\mathcal{L}}

\global\long\def\mF{\mathcal{F}}

\global\long\def\mA{\mathcal{A}}

\global\long\def\mH{\mathcal{H}}

\global\long\def\mX{\mathcal{X}}

\global\long\def\dist{d}

\global\long\def\HX{\entro\left(X\right)}
 \global\long\def\entropyX{\HX}

\global\long\def\HY{\entro\left(Y\right)}
 \global\long\def\entropyY{\HY}

\global\long\def\HXY{\entro\left(X,Y\right)}
 \global\long\def\entropyXY{\HXY}

\global\long\def\mutualXY{\mutual\left(X;Y\right)}
 \global\long\def\mutinfoXY{W\mutualXY}

\global\long\def\given{\mid}

\global\long\def\gv{\given}

\global\long\def\goto{\rightarrow}

\global\long\def\asgoto{\stackrel{a.s.}{\longrightarrow}}

\global\long\def\pgoto{\stackrel{p}{\longrightarrow}}

\global\long\def\dgoto{\stackrel{d}{\longrightarrow}}

\global\long\def\ll{\mathit{l}}

\global\long\def\logll{\mathcal{L}}

\global\long\def\zb{\boldsymbol{z}}
 \global\long\def\wb{\boldsymbol{w}}
 \global\long\def\xb{\boldsymbol{x}}
 \global\long\def\fb{\boldsymbol{f}}

\global\long\def\bzero{\vt0}

\global\long\def\bone{\mathbf{1}}

\global\long\def\bff{\vt f}

\global\long\def\bx{\boldsymbol{x}}

\global\long\def\bX{\boldsymbol{X}}

\global\long\def\bW{\mathbf{W}}

\global\long\def\bH{\mathbf{H}}

\global\long\def\bL{\mathbf{L}}

\global\long\def\tbx{\tilde{\bx}}

\global\long\def\by{\boldsymbol{y}}

\global\long\def\bY{\boldsymbol{Y}}

\global\long\def\bz{\boldsymbol{z}}

\global\long\def\bZ{\boldsymbol{Z}}

\global\long\def\bu{\boldsymbol{u}}

\global\long\def\bU{\boldsymbol{U}}

\global\long\def\bv{\boldsymbol{v}}

\global\long\def\bV{\boldsymbol{V}}

\global\long\def\bw{\vt w}

\global\long\def\balpha{\gvt\alpha}

\global\long\def\bbeta{\gvt\beta}

\global\long\def\bmu{\gvt\mu}

\global\long\def\btheta{\boldsymbol{\theta}}
 \global\long\def\thetab{\boldsymbol{\theta}}

\global\long\def\blambda{\boldsymbol{\lambda}}
 \global\long\def\lambdab{\boldsymbol{\lambda}}

\global\long\def\realset{\mathbb{R}}

\global\long\def\realn{\real^{n}}

\global\long\def\natset{\integerset}

\global\long\def\interger{\integerset}

\global\long\def\integerset{\mathbb{Z}}

\global\long\def\natn{\natset^{n}}

\global\long\def\rational{\mathbb{Q}}

\global\long\def\realPlusn{\mathbb{R_{+}^{n}}}

\global\long\def\comp{\complexset}
 \global\long\def\complexset{\mathbb{C}}

\global\long\def\and{\cap}

\global\long\def\compn{\comp^{n}}

\global\long\def\comb#1#2{\left({#1\atop #2}\right) }

\section{Introduction \label{sec:intro}}

Every baby is expected at full term. However, still 10-15\% of all
infants will be born before 37 weeks as preemies \cite{barros2015distribution}.
Preterm birth is a major cause of infant mortality, developmental
retardation and long-term disabilities \cite{vovsha2014predicting}.
The earlier the arrival, the longer the baby stays in intensive care,
causing more cost and stress for the mother and the family. Predicting
preterm births is highly critical as it would guide care and early
interventions. 

Most existing research on preterm birth prediction focuses on identifying
individual risk factors in the hypothesis-testing paradigm under highly
controlled settings \cite{mercer1996preterm}. The strongest predictor
has been prior preterm births. But this does not apply for first-time
mothers or those without prior preterm births. There are few predictive
systems out there, but the predictive power is very limited. One of
the best known studies, for example, achieved only sensitivity of
24.2\% at specificity of 28.6\% for first-time mothers \cite{mercer1996preterm}.
Machine learning techniques have been used with promising results
\cite{goodwin2001data,vovsha2014predicting}. For example, an Area
Under the ROC curve (AUC) of 0.72 was obtained in \cite{goodwin2001data}
using a large observational dataset.

This paper asks the following questions: can we learn to predict preterm
births from a large observational database without going through the
hypothesis testing phase? What is the best way to generate and combine
hypotheses from data? To this end, this work differs from previous
clinical research by first generating hundreds (or even thousands)
of potential signals and then developing machine learning methods
that can handle many irrelevant features. Our goal is \emph{to} \emph{develop
a method that: (i) derives a compact set of risk factors with quantified
uncertainties; (ii) estimates the preterm risks; and (iii) explains
the prediction made.}

In other words, we derive a prediction rule to be used in practice.
This demands \emph{interpretability} \cite{freitas2014comprehensible,ruping2006learning}
and \emph{stability} \cite{yu2013stability}. While interpretability
is self-explanatory \cite{ruping2006learning}, stability refers to
the model that is stable under data resampling \cite{tran_et_al_kais14,yu2013stability}
(i.e., model parameters do not change significantly when re-estimated
from a new data sample). Thus stability is necessary for reproducibility
and thus must also be enforced. 

Our approach is based on $\ell_{1}$-penalized logistic regression
\cite{meier2008group}, which is stabilized by a graph of feature
correlations \cite{tran_et_al_kais14}, resulting in a model called
Stabilized Sparse Logistic Regression (SSLR). Bootstrap is then utilized
to estimate the mode of model posterior as well as to compute feature
importance. The prediction rule is then derived by keeping only top
$k$ most important features whose weights are scaled and rounded
to integers. For estimating the upper-bound of prediction accuracy,
we derive a sophisticated ensemble classifier called Randomized Gradient
Boosting (RGB) by combining powerful properties of Random Forests
\cite{breiman2001random} and Stochastic Gradient Boosting \cite{friedman2002stochastic}.

The models are trained and validated on a large observational database
consisting of 15,814 women and 18,836 pregnancy episodes. The SSLR
achieves AUCs of 0.85 and 0.79 for 34-week and 37-week preterm birth
predictions, respectively, only slightly lower than those by RGB (0.86
and 0.81). The results are better than a previous study with matched
size and complexity \cite{goodwin2001data} (AUC 0.72). A simplified
10-item prediction rule suffers only a small loss in accuracy (AUCs
0.84 and 0.77 for 34 and 37 week prediction, respectively) but has
much better transparency and interpretability.

\section{Cohort}

\begin{table}
\begin{centering}
\begin{tabular}{|l|r|r|}
\hline 
\# episodes  & \multicolumn{2}{r|}{18,836}\tabularnewline
\hline 
\# mothers & \multicolumn{2}{r|}{15,814}\tabularnewline
\hline 
\# multifetal epis. & \multicolumn{2}{r|}{500 (2.7\%)}\tabularnewline
\hline 
Age  & \multicolumn{2}{r|}{32.1 (STD: 4.9)}\tabularnewline
\hline 
\emph{\#preterm births}: & \emph{\textless{}37 weeks} & \emph{\textless{}34 weeks}\tabularnewline
\hline 
\textendash total & 2,067 (11.0\%) & 1,177 (6.3\%)\tabularnewline
\textendash spontaneous  & 1,283 (62.1\%)  & 742 (63.0\%)\tabularnewline
\textendash elective & 754 (36.5\%) & 436 (37.0\%)\tabularnewline
\hline 
\end{tabular}
\par\end{centering}
\caption{Data statistics.\label{tab:Data-statistics}}
\end{table}

\subsection{Cohort Selection \label{subsec:Cohort-Selection}}

We acquired a large observational dataset from Royal North Shore (RNS)
hospital located in NSW, Australia. The RNS has 18,836 pregnancy episodes
collected during the preriod of 2007-2015. The data statistics are
summarized in Table~\ref{tab:Data-statistics}. 

\paragraph{Preterm Birth Definition}

A birth is considered preterm if it occurs before 37 full weeks of
gestation. There are two types of preterm births \textendash{} spontaneous
and elective (or indicated). Spontaneous births occur naturally without
clinical interventions, and this accounts for about 60-70\% of all
preterm births. In our database, the rules for figuring out spontaneous
and elective, as specified by our senior clinician, are:
\begin{itemize}
\item \textsf{Spontaneous}: \emph{PPROM} $\vee$ (\emph{Preterm}$\wedge$
\emph{UterineActiv.Adm.} $\wedge$ $\neg$\emph{PrelaborInterv.}) 
\item \textsf{Elective}: \emph{Preterm} $\wedge$ $\neg$\emph{PPROM} $\wedge$
$\neg$\emph{UterineActiv.Adm}. $\wedge$ (\emph{PrelaborInterv.}
$\vee$ \emph{CaesareanSect.})
\end{itemize}
where PPROM stands for Preterm Premature Rupture of Membranes. This
may leave a small portion of births not falling into either category
($\sim1.5\%$). 

\paragraph{Sub-cohort Selection}

\begin{figure*}
\begin{centering}
\includegraphics[width=0.32\textwidth,height=0.32\textwidth]{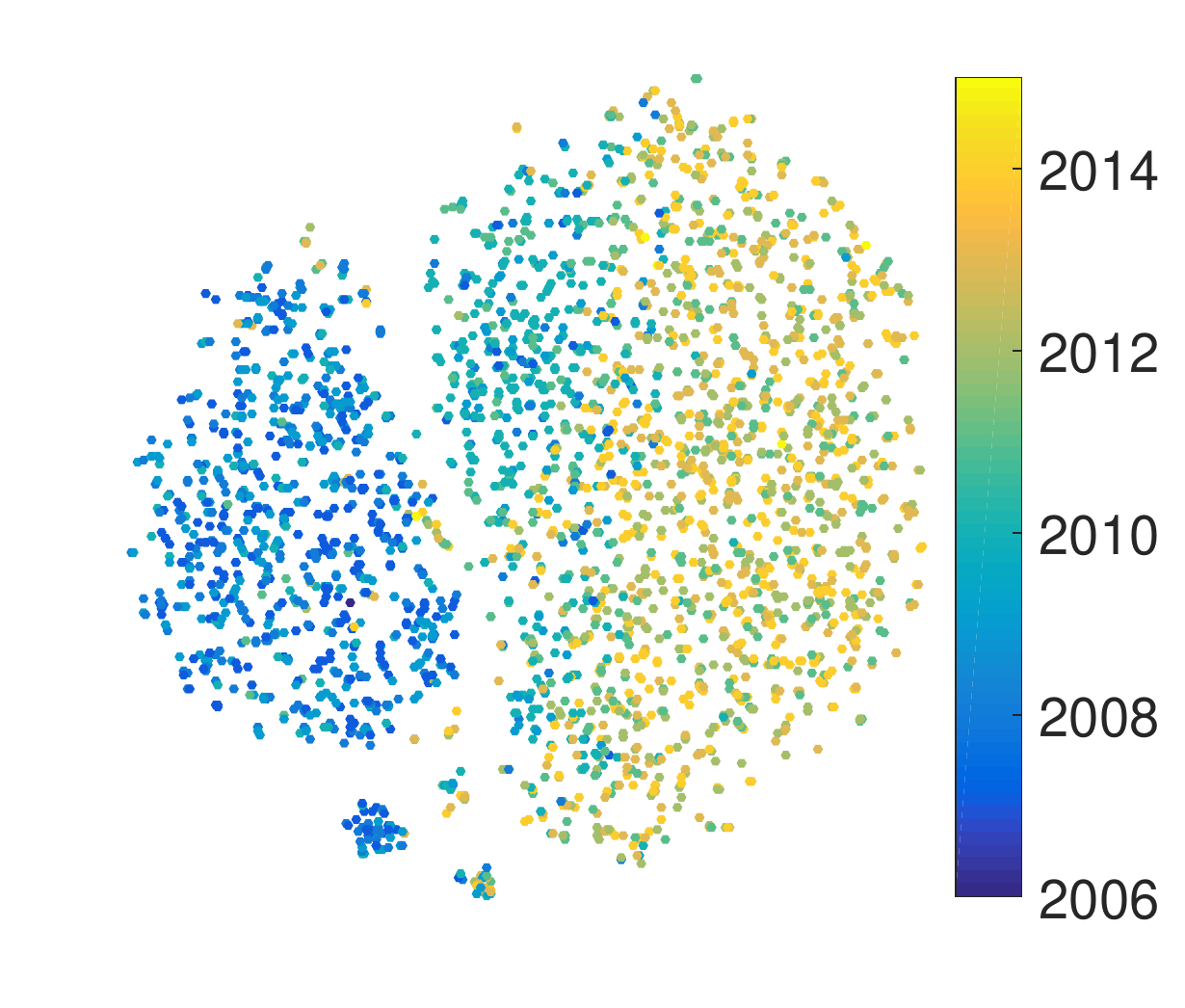}\includegraphics[width=0.32\textwidth,height=0.32\textwidth]{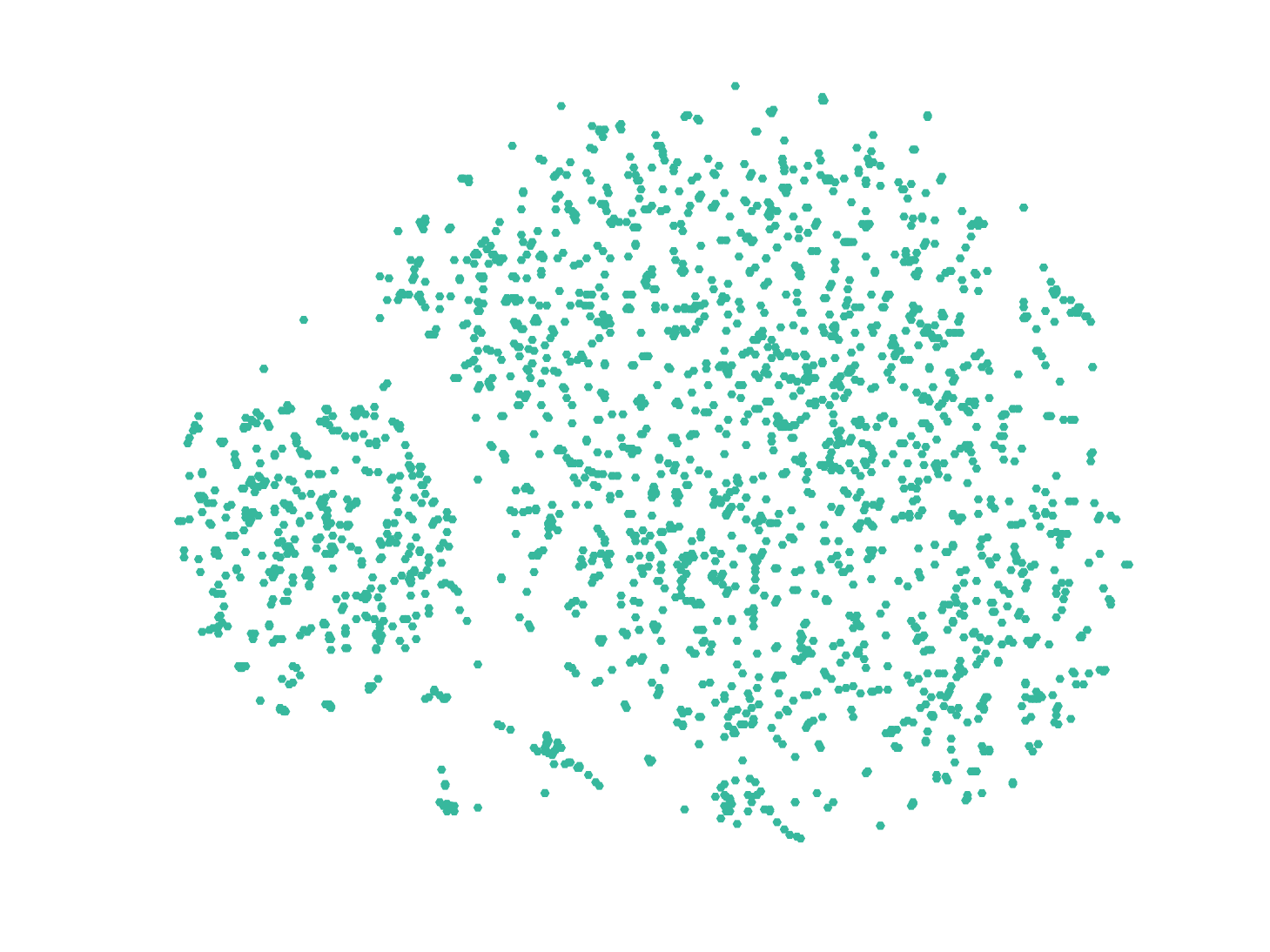}\includegraphics[width=0.32\textwidth,height=0.32\textwidth]{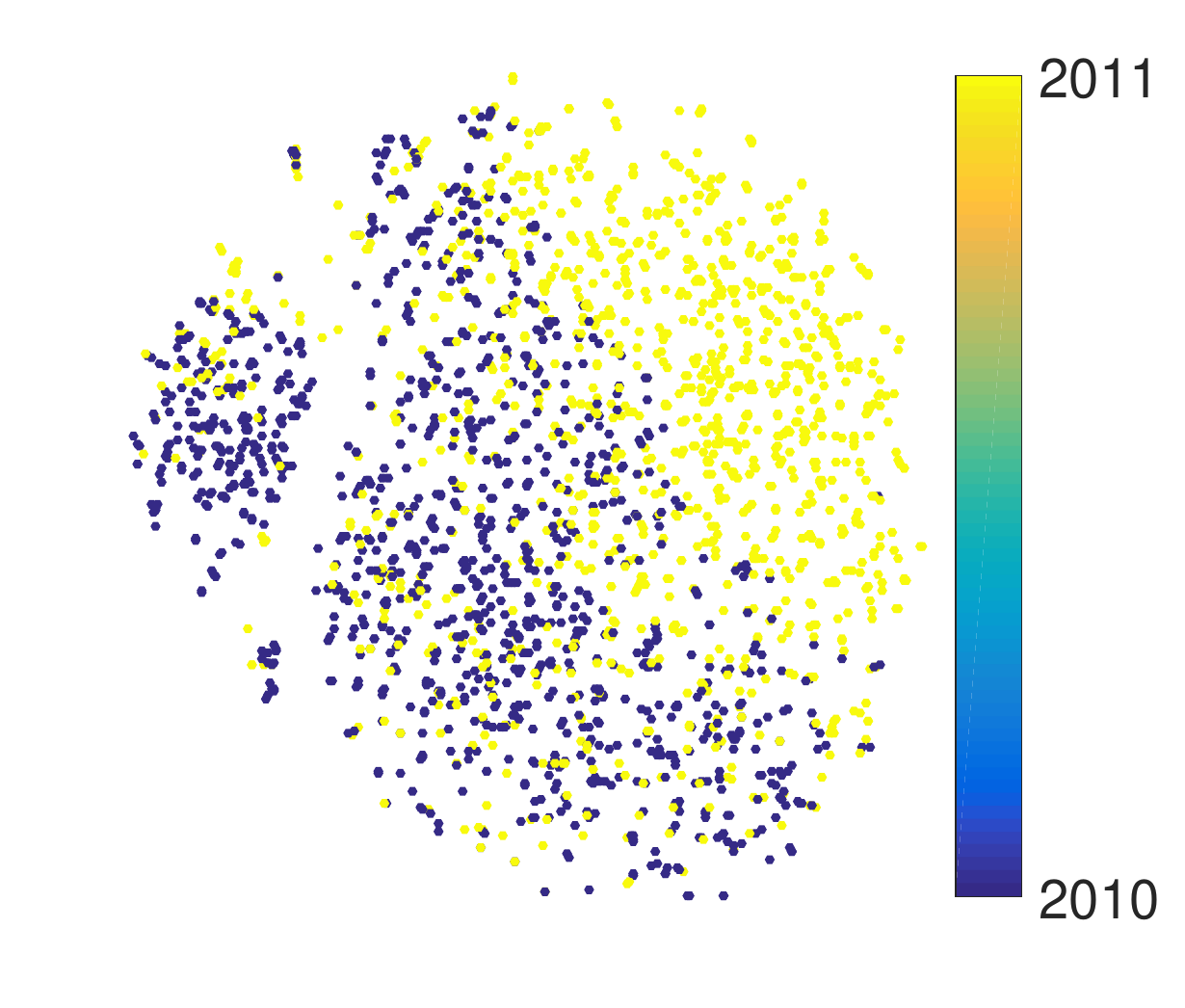}
\par\end{centering}
\caption{Covariate shifting over time in the RNS dataset (Left: 2006-2015;
Middle: 2010, Right: 2010-2011). Each point is a pregnancy episode.
Colors represent years of birth. Best viewed in colors.\label{fig:Covariate-shifting}}
\end{figure*}

Our initial analysis revealed a critical, undocumented issue: \emph{covariate
shifting} over time, probably due to the evolution of data collection
protocol. Using features extracted in the next subsection, we embedded
data into 2D using t-SNE \cite{van2008visualizing} and visually examined
the shifting. Fig.~\ref{fig:Covariate-shifting} (left) plots data
points (episodes) coded in colors which correspond to the year the
episodes occurred. The colors change gradually from left (2006, dark
blue) to right (2015, bright yellow) with a sudden discontinuity in
the middle of 2010. Fig.~\ref{fig:Covariate-shifting} (middle) amplifies
the year 2010. There is also a big shift from 2010 to 2011, as shown
in Fig.~\ref{fig:Covariate-shifting} (right). Years 2011-2015 are
more uniformly distributed. For this reason, we will work exclusively
on the data collected between 2011-2015 as they are more recent.

\subsection{Feature Generation}

We operate under a hypothesis-free mode \textendash{} the data is
collected in hospital operations. The biggest challenge in generating
features from an observational database is to detect and prevent the
so-called ``leakage'' problem \cite{kaufman2012leakage}. This happens
when recorded information implicitly indicates the outcome to be predicted.
In pregnancy databases, it could be procedures and tests that are
performed late in the course of pregnancy. When it happens, it may
already indicate a full-term birth. With our clinicians and local
database experts, we verify the generated feature list to prevent
this from happening. We explicitly extracted features that occurred
before week 25 of gestation. 

During a pregnancy, a woman typically visits the hospital several
times before labor. At 20-25 weeks, it is critical to estimate the
risks, one of which is the risk of preterm birth. A care model may
be decided by clinician depending on the perceived risks. The care
allocation contains important information not present in other risk
factors, and thus we will treat them separately. A part from real-valued
measurements (such as BMI and weights), many features are discrete,
for example, a medication name under the data field ``medication''.
Thus discrete features are counts of such discrete symbols. The database
also contains a certain amount of free text, documenting the findings
by clinicians. For each text field, we extracted unigrams after removing
stop words. For robustness, rare features that occurs in less than
1\% of data points were removed. Finally, we retained 762 features
without textual information, and 2,770 features with textual information.

\section{Methods \label{sec:methods}}

\subsection{Stabilized Sparse Logistic Regression \label{subsec:Stabilizing-sparse-models}}

For a feature vector $\xb\in\mathbb{R}^{p}$, we focus on building
a linear prediction rule: $f(\xb)=w_{0}+\sum_{i=1}^{p}w_{i}x_{i}$,
where $w_{0},w_{1},...,w_{p}$ are feature weights. For binary classification,
we use logistic regression to define the probability of outcome: $P(y=1\mid\xb)=1/\left(1+e^{-f(\xb)}\right)$.
To work with a large number of features, we derive a sparse solution
by minimizing the following $\ell_{1}$-penalized loss \cite{meier2008group}:
\begin{equation}
L=-\sum_{d=1}^{n}\log P(y^{(d)}\mid\xb^{(d)})+\lambda\sum_{i}|w_{i}|\label{eq:lasso}
\end{equation}
where $d$ denotes data points, and $\lambda>0$ is the penalty factor
responsible for driving weights of weak features toward zeros.

However, sparsity invites instability, that is, the selected features
vary greatly if we slightly perturb the dataset \cite{gopakumar2014bstabilizing,tran_et_al_kais14,zou2005regularization}.
One reason is that when two features are highly correlated, the $\ell_{1}$-penalty
will pick one randomly. In healthcare practice, multiple processes
and data views are often recorded at the same time, causing a high
level of redundancy in observational data. This results in \emph{multiple
models (and feature subsets) of equal predictive performance}. This
behavior is undesirable because the derived models are unstable to
earn trust from clinicians.

An effective way to reduce instability under correlated data is introduced
in \cite{tran_et_al_kais14}. More precisely, we assert that correlated
features will have similar weights. This is realized through the following
objective function:
\begin{equation}
L=L_{0}+\lambda\sum_{i}\left(\alpha|w_{i}|+\frac{1-\alpha}{2}\left(w_{i}-\sum_{j\ne i}S_{ij}w_{j}\right)^{2}\right)\label{eq:random-walk}
\end{equation}
where $L_{0}=-\sum_{d=1}^{n}\log P(y^{(d)}\mid\xb^{(d)})$, $\alpha\in[0,1]$,
and $S_{ij}>0$ is the similarity between features $i$ and $j$,
subject to $\sum_{j\ne i}S_{ij}=1$. This feature-similarity regularizer
is equivalent to a multivariate Gaussian prior of mean $\mathbf{0}$
and precision matrix $\left(1-\alpha\right)\left(\mathbf{S}-\mathbf{I}\right)^{\top}\left(\mathbf{S}-\mathbf{I}\right)$
where $\mathbf{I}$ is the identity matrix. Hence minimizing the loss
in Eq.~(\ref{eq:random-walk}) is to find the \emph{maximum a posterior}
(MAP), where the prior is a product of a Laplace and a Gaussian distributions.
In this paper, the similarity matrix $\mathbf{S}$ is computed using
the cosine between data columns (each corresponding to a feature).
We will refer to this model as Stabilized Sparse Logistic Regression
(SSLR).

\subsection{Deriving Prediction Rule and Risk Curve\label{subsec:Deriving-prediction-rule}}

The prediction rule and risk curve are generated using the following
algorithm.\textbf{ }Given number of retained features $k$ and number
of bootstraps $B$, the steps are as follows:
\begin{enumerate}
\item \textbf{Bootstrap model averaging}: $B$ SSLR models (Sec.~\ref{subsec:Stabilizing-sparse-models})
are estimated on $B$ data bootstraps. The feature weights are then
averaged. Together with the MAP estimator in Eq.~(\ref{eq:random-walk}),
this model averaging is closely related to finding a mode of the parameter
posterior in an approximate Bayesian setting. This procedure is expected
to further improve model stability by simulating data variations \cite{wang2011random}.
\item \textbf{Feature selection}: Features are ranked based on importance,
which is averaged weight $\times$ feature standard-deviation \cite{friedman2008predictive}.
This measure of feature importance is insensitive to feature scale,
and encodes the feature strength, stability and entropy. Top $k$
features are kept.
\item \textbf{Prediction rule construction}: Weights of selected features
are linearly transformed and rounded to sensible integers. For example,
the weights range from 1 to 10 for positive weights, and from -10
to -1 for negative weights. The prediction rule has the following
form: $f^{*}(\xb)=\sum_{j=1}^{k}\eta_{j}x_{j}$ where $\eta_{j}$
are non-zero integers. We shall refer to features with positive weights
as \emph{risk factors}, and those with negative weigths as \emph{protective
factors}.
\item \textbf{Risk curve}: The prediction rule is then used to score all
patients. The scores are converted into risk probability using univariate
logistic regression. This produces a risk curve.
\end{enumerate}

\subsection{Randomized Gradient Boosting \label{subsec:Randomized-gradient-boosting}}

State-of-the-art classifiers are often ensembles such as Random Forests
(RF) \cite{breiman2001random} and Stochastic Gradient Boosting (SGB)
\cite{friedman2002stochastic}. To test how simplified sparse linear
methods may fare against complex ensembles, we develop a hybrid RF/SGB
called Randomized Gradient Boosting (RGB) that estimates the outcome
probability as follows:
\[
P(y=1\mid\xb)=\frac{1}{1+\exp\left(-\sum_{t}\beta_{t}h_{t}(\xb_{t})\right)}
\]
where $\beta_{t}\in(0,1)$ is a small learning rate, $\xb_{t}\in\xb$
is a feature subset, and each $h_{t}(\xb_{t})$ is a regression tree,
which is added in a sequential manner as in \cite{friedman2002stochastic}.
Following \cite{breiman2001random}, and each non-terminal node is
split based on a small random subset of features.

\section{Results \label{sec:Experiments}}

\subsection{Evaluation Approach/Study Design}

The data is randomly spitted into two parts: 2/3 for training and
1/3 for testing. To be consistent with the practice of clinical research,
for the test data \emph{we maintain balanced classes} through under-sampling
of the majority class. The parameters for Stabilized Sparse Logistic
Regression (SSLR) in Eq.~(\ref{eq:random-walk}) are set as $\alpha=0.5$,
and $\lambda=5$. The Randomized Gradient Boosting (RGB, Sec.~\ref{subsec:Randomized-gradient-boosting})
had 500 decision trees learnt from a learning rate of 0.03, and each
tree had 256 leaves at most. Each tree is trained on a random subset
of $m=\left\lfloor p/3\right\rfloor $ features, and each node split
is based on a random sub-subset of $\left\lceil m/3\right\rceil $
features.

For performance measures, we report sensitivity (recall), specificity,
NPV, PPV (precision), F-measure (2$\times$recall$\times$precision/(recall
+ precision)) and AUC. Except for AUC, the other measures depend on
the decision threshold at which the prediction is made, that is we
predict $\hat{y}=1$ if $P(y=1\mid\xb)\ge\tau$ for threshold $\tau\in(0,1)$.
We chose the threshold so that sensitivity matches specificity in
the training data. 

\subsection{Visual Examination}

\begin{figure}
\vspace{-5mm}
\begin{tabular}{>{\centering}p{0.5\textwidth}>{\centering}p{0.5\textwidth}}
\subfloat[Distribution of term/preterm births. Each point is an episode. Bright
color represent preterm births (less than 37 full weeks of gestation).
Best viewed in colors.\label{fig:Distribution-of-preterms}]{\includegraphics[width=0.45\textwidth,height=0.45\textwidth]{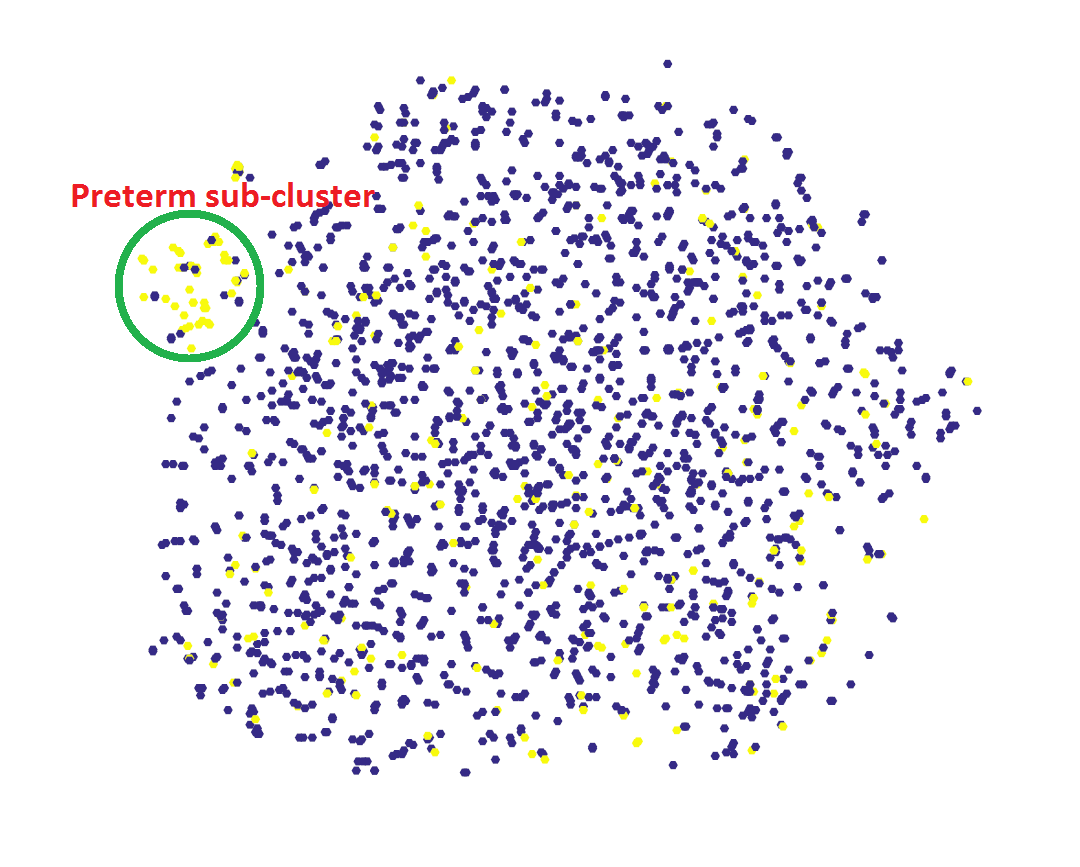}

} & \subfloat[Risk curve for prediction rule in Table~\ref{tab:Prediction-rule-with-care}.\label{fig:Risk-curve}]{\includegraphics[width=0.45\textwidth,height=0.45\textwidth]{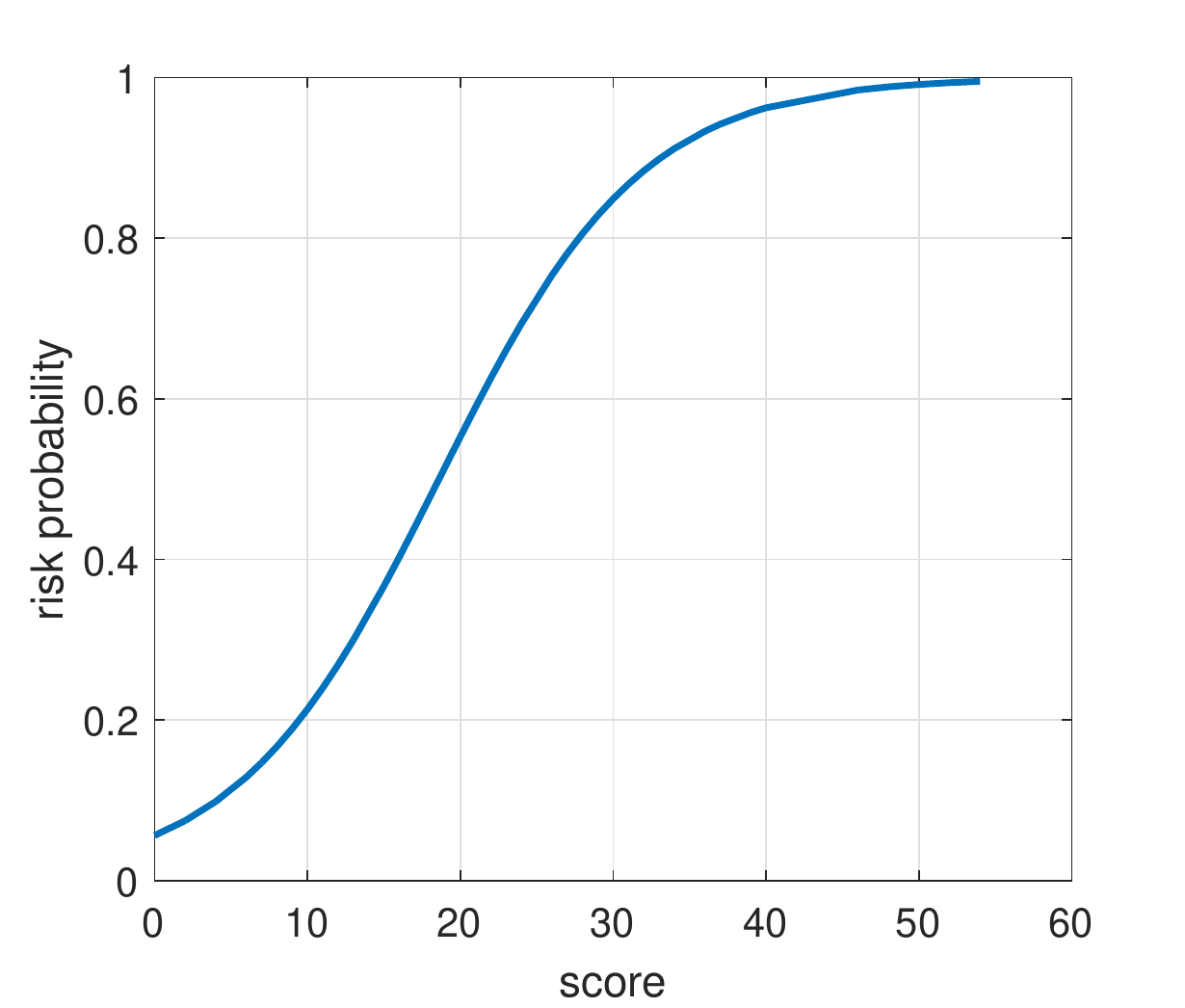}}\tabularnewline
\end{tabular}

\caption{Preterm birth distribution and estimated risk curve.}
\vspace{-5mm}
\end{figure}

To visually examine the difficulty of the prediction problem, we embed
data points (episodes) into 2D using t-SNE \cite{van2008visualizing}.
Fig.~\ref{fig:Distribution-of-preterms} plots data points coded
in colors corresponding to preterm or full-term. There is a small
cluster mostly consisting of preterm births, and a big cluster in
which preterm births are randomly mixed with term births. This suggests
that there are no simple linear hyper-planes that can separate the
preterm births from the rest.

\subsection{Prediction Results}

\begin{table}
\begin{centering}
\vspace{-5mm}
\begin{tabular}{|l|c|c|c|c|c|c|}
\hline 
 & \multicolumn{2}{c|}{\textbf{Obs}} & \multicolumn{2}{c|}{\textbf{Obs+care}} & \multicolumn{2}{c|}{\textbf{Obs+care+text}}\tabularnewline
\hline 
 & SSLR & RGB & SSLR & RGB & SSLR & RGB\tabularnewline
\hline 
\hline 
Sensitivity & 0.723 & 0.621 & 0.734 &  0.644 & 0.698 & 0.720\tabularnewline
\hline 
Specificity & 0.643 & 0.820 & 0.711 & 0.841 & 0.732 & 0.740\tabularnewline
\hline 
NPV & 0.690 & 0.675 & 0.719 & 0.693 & 0.699 & 0.717\tabularnewline
\hline 
PPV & 0.679 & 0.783 & 0.726 & 0.809 & 0.731 & 0.743\tabularnewline
\hline 
F-measure & 0.700 & 0.693 & 0.730 & 0.717 & 0.714 & 0.732\tabularnewline
\hline 
\end{tabular}
\par\end{centering}
\caption{Classifier performance for 37-week preterm births. Obs = observed
features without care allocation. SSLR = Stabilized Sparse Logistic
Regression, RGB = Randomized Gradient Boosting.\label{tab:perf-37wks-2011+}\vspace{-5mm}
}
\end{table}

We investigate multiple settings: observed features only (Obs), observation
with booking \& care allocation (Obs+care), and observation with textual
information (Obs+care+text). Table.~\ref{tab:perf-37wks-2011+} reports
sensitivity, specificity, NPV, PPV, and F-measure by SSLR and RGB.
The sensitivity for SSLR ranges from 0.698\textendash 0.734 at specificity
of 0.643\textendash 0.732. The sensitivity for RGB is between 0.621\textendash 0.720
at specificity of 0.740\textendash 0.841. The F-measures for both
classifiers are comparable in the range of 0.693\textendash 0.732.

\begin{table}
\begin{centering}
\begin{tabular}{|l|l|c|c|c|}
\hline 
\textbf{Outcome} & \textbf{Algo.} & \textbf{Obs} & \textbf{Obs+care} & \textbf{Obs+care+text}\tabularnewline
\hline 
\hline 
\multirow{2}{*}{Spontaneous} & SSLR & 0.717 & 0.744 & 0.754\tabularnewline
\cline{2-5} 
 & RGB & 0.750 & 0.761 & 0.773\tabularnewline
\hline 
\multirow{2}{*}{All} & SSLR & 0.764 & 0.791 & 0.790\tabularnewline
\cline{2-5} 
 & RGB & 0.782 & 0.804 & 0.807\tabularnewline
\hline 
\end{tabular}
\par\end{centering}
\caption{AUC for 37-week preterm. See Tab.~\ref{tab:perf-37wks-2011+} for
legend explanation. \label{tab:AUC-37wks-2011+}}
\end{table}

Table~\ref{tab:AUC-37wks-2011+} reports the AUC for different settings
(for spontaneous births only and all cases). For spontaneous births,
the highest AUC of 0.773 is achieved by RGB using all available information.
For all births, the highest AUC is 0.807, also by RGB with all information.
Overall, RGB fares slightly better than SSLR in AUC measure. Care
information, such as booking and allocation of care, has a good predictive
power. 

Table~\ref{tab:AUC-34wks-2011+} reports the AUC for 34-week prediction.
For spontaneous births prediction, the largest AUC of 0.849 is achieved
by RGB using care information. For both elective and spontaneous births,
the largest AUC is 0.862.

\begin{table}
\begin{centering}
\begin{tabular}{|l|l|c|c|}
\hline 
\textbf{Outcome} & \textbf{Algo.} & \textbf{Obs} & \textbf{Obs+care}\tabularnewline
\hline 
\hline 
\multirow{2}{*}{Spontaneous} & SSLR & 0.806 & 0.828\tabularnewline
\cline{2-4} 
 & RGB & 0.841 & 0.849\tabularnewline
\hline 
\multirow{2}{*}{All} & SSLR & 0.834 & 0.850\tabularnewline
\cline{2-4} 
 & RGB & 0.857 & 0.862\tabularnewline
\hline 
\end{tabular}
\par\end{centering}
\caption{AUC for 34-week preterm. Tab.~\ref{tab:perf-37wks-2011+} for legend
explanation. \label{tab:AUC-34wks-2011+}}
\end{table}

\subsection{Prediction Rules}

\begin{table}
\begin{centering}
\begin{tabular}{|l|c|c|c|c|}
\hline 
\textbf{Outcome} & \textbf{weeks} & \textbf{Risk factors only } & \textbf{W/Protect. factors} & \textbf{SSLR}\tabularnewline
\hline 
\hline 
\multirow{2}{*}{Spontaneous} & 34 & 0.804 & 0.823 & \emph{0.828}\tabularnewline
\cline{2-5} 
 & 37 & 0.725 & 0.728 & \emph{0.743}\tabularnewline
\hline 
\multirow{2}{*}{Elective/spontaneous} & 34 & 0.816 & 0.837 & \emph{0.850}\tabularnewline
\cline{2-5} 
 & 37 & 0.757 & 0.767 & \emph{0.784}\tabularnewline
\hline 
\end{tabular}
\par\end{centering}
\caption{AUC of prediction rules with care information. The left column is
the SSLR with all factors for reference. Risk factors are those with
positive weights, whereas protective factors have negative weights.
\label{tab:Prediction-rule-performance-with-care}}
\end{table}

Prediction rules are generated using the procedure described in Sec.~\ref{subsec:Deriving-prediction-rule}.
Table~\ref{tab:Prediction-rule-performance-with-care} reports the
predictive performance of generated rules with 10 items. Generally
the performance drops by several percent points. Using protective
factors (those with negative weights) is better, suggesting that they
should be used rather than discarded. Table~\ref{tab:Prediction-rule-no-care}
lists the items and their associated weights (with standard deviations)
for the case \emph{without} care information. The top three risk factors
are \emph{multiple fetuses}, \emph{cervix incompetence} and \emph{prior
preterm births}. Other risk factors include domestic violence, history
of hypertension, illegal use of marijuana, diabetes history and smoking.
Likewise, Table~\ref{tab:Prediction-rule-with-care} reports for
the case \emph{with} care information, which plays an important roles
as risk factors.

\begin{table}
\begin{centering}
\begin{tabular}{|l|c|}
\hline 
\textbf{Risk factor } & \textbf{Score (}$\pm$\textbf{Std)}\tabularnewline
\hline 
1. Number of fetuses at 20 weeks \textgreater{}= 2  & 10 ($\pm$0.7)\tabularnewline
2. Cervix shortens/dilates before 25wks  & 8 ($\pm$1.3)\tabularnewline
3. Preterm pregnancy  & 3 ($\pm$0.7)\tabularnewline
4. Domestic violence response: deferred & 2 ($\pm$0.8)\tabularnewline
5. Hist. Hypertension: essential & 2 ($\pm$1.2)\tabularnewline
6. Illegal drug use: Marijuana & 2 ($\pm$1.0)\tabularnewline
\hline 
7a. Hist. of Diabetes Type 1 & 2 ($\pm$1.0)\tabularnewline
8a. Daily Cigarette: one or more & 2 ($\pm$0.9)\tabularnewline
9a. Prescription 1st Trimester: insulin & 1 ($\pm$0.8)\tabularnewline
10a. Baby Aboriginal Or Tsi: yes  & 1 ($\pm$0.7)\tabularnewline
\hline 
7b. Ipc Gen. Confident: sometimes & -2 ($\pm$0.7)\tabularnewline
8b. Ultrasound Indication :other & -2 ($\pm$0.8)\tabularnewline
9b. Ipc Emotional Support: yes & -3 ($\pm$0.5)\tabularnewline
10b. Ipc Generally Confident: yes & -3 ($\pm$0.5)\tabularnewline
\hline 
\end{tabular} 
\par\end{centering}
\caption{10-item prediction rule, \emph{without} care information. (The first
rule with items {[}1-6,7a-10a{]} (risk factors only) achieves AUC
0.702; the second rule with items {[}1-6; 7b-10b{]} (risk+protective
factors) achieves AUC 0.743).\label{tab:Prediction-rule-no-care}}
\end{table}

\begin{table}
\begin{centering}
\begin{tabular}{|l|c|}
\hline 
\textbf{Risk factor } & \textbf{Score (}$\pm$\textbf{Std) }\tabularnewline
\hline 
1. Number of fetuses at 20 weeks \textgreater{}= 2  & 10 ($\pm$0.9)\tabularnewline
2. Cervix shortens/dilates before 25wks  & 9 ($\pm$1.7)\tabularnewline
3. Allocated Care: private obstetrician & 6 ($\pm$0.8)\tabularnewline
4. Booking Midwife: completed birth & 5 ($\pm$1.2)\tabularnewline
5. Allocated Care: hospital based  & 4 ($\pm$0.6)\tabularnewline
6. Preterm pregnancy  & 3 ($\pm$0.8)\tabularnewline
7. Illegal drug use: Marijuana & 2 ($\pm$0.9)\tabularnewline
8. Hist. Hypertension: essential & 2 ($\pm$1.4)\tabularnewline
\hline 
9a. Dv Response: deferred & 2 ($\pm$1.2)\tabularnewline
10a. Daily Cigarette: one or more & 1 ($\pm$1.1)\tabularnewline
\hline 
9b. Ipc Emotional Support: yes & -3 ($\pm$0.6)\tabularnewline
10b. Ipc Generally Confident: yes & -3 ($\pm$0.5)\tabularnewline
\hline 
\end{tabular}
\par\end{centering}
\caption{10-item prediction rule, \emph{with} care information. Dv: domestic
violence. The first rule with items {[}1-8,9a-10a{]} (risk factors
only) achieves AUC 0.757; the second rule with items {[}1-6; 7b-10b{]}
(risk+protective factors) achieves AUC 0.767.\label{tab:Prediction-rule-with-care}}
\end{table}

Fig.~\ref{fig:Risk-curve} shows the risk curve estimated from the
first prediction rule in Table~\ref{tab:Prediction-rule-with-care}
(without protective factors). When the score is 0, there is still
a 5.3\% chance of preterm. That says that the risk factors here only
account for 50\% of preterm births. When the score is 10 (e.g., with
twins), the risk doubles.

\section{Discussion and Related Work\label{sec:Discussion}}

We have presented methods for predicting preterm births from high-dimensional
observational databases. The methods include: (i) discovering and
quantifying risk factors, and (ii) deriving simple, interpretable
prediction rules. The main methodological novelties are (a) the use
of stabilized sparse logistic regressions (SSLR) for deriving stable
linear prediction models, and (b) the use of bootstrap model averaging
for distill simple prediction rules in an approximate Bayesian fashion.
To estimated the upper-bound of model accuracy for given data, we
also introduced Randomized Gradient Boosting, which is a hybrid of
Random Forests \cite{breiman2001random} and Stochastic Gradient Boosting
\cite{friedman2002stochastic}. 

\subsubsection*{Findings}

For 37-week preterm births, the highest AUC using RGB is in the range
0.78 using only observational information, and in the range 0.80-0.81
with care information (booking + allocation decision). The SSLR is
slightly worse with the AUC in the range of 0.76 with only observational
information, and AUC of 0.79 with care decision.Thus, care information
has a good predictive power. This is expected since it encodes doctor's
knowledge in risk assessment. It is also likely to be available later
in the course of pregnancy. The results are better than a previous
study with matched size and complexity \cite{goodwin2001data} (AUC
= 0.72). Simplified prediction rules with only 10 items suffer some
small loss in accuracy. The AUCs are 0.74 and 0.77 with and without
care information, respectively. The payback is much better in transparency
and interpretability.

\subsubsection*{Related Work}

\emph{Preterm birth prediction} has been studied for several decades
\cite{de2005prediction,goldenberg1998preterm,iams2001preterm,macones1999prediction,vovsha2014predicting}.
Most existing research either focuses on deriving individual predictive
factors, or builds prediction model under highly controlled data collection.
Three most common known risk factors are: prior preterm births, cervical
incompetence and multiple fetuses. These agree with our findings (e.g.,
Table~\ref{tab:Prediction-rule-no-care}). Data mining approaches
that leverage observational databases have been attempted in \cite{goodwin2001data}
and \cite{vovsha2014predicting} showing a great promise. 

\emph{Clinical prediction rules} have been widely used in practice
\cite{gage2001validation}. A popular approach is logistic regression
with scaled and rounded coefficients. \emph{Model stability} has been
studied in biomedical prediction \cite{austin2004automated,he2010stable,gopakumar2014bstabilizing,tran_et_al_kais14}.
The machine learning community has worked and commented on \emph{interpretable
prediction rules} in multiple places \cite{bien2011prototype,carrizosa2016strongly,emad2015semiquantitative,freitas2014comprehensible,huysmans2011empirical,ruping2006learning,vellido2012making,wang2015trading}.
There are been applications to biomedical domains \cite{haury2011influence,song2013random,ustun2015supersparse}.
In \cite{ustun2015supersparse}, the authors seek to derive a sparse
linear integer model (SLIM) where the coefficients are linear. The
simplification of complex models is also known as \emph{model distillation
}\cite{hinton2015distilling} or model compression \cite{bucilua2006model}.
Most current work in model distillation focuses on deep neural networks,
which are hard to interpret. 

\emph{}

\subsubsection*{Limitations}

The models derived in this paper are subject to the quality of data
collected. For example, the covariate shift problem can occur within
a hospital over time, as pointed out in Sec.~\ref{subsec:Cohort-Selection}.
This study is also limited to data collected just for the pregnancy
visits. There may be more predictive information in the electronic
medical records. However, initial inquiry revealed that since pregnant
women are relatively young, the medical records are rather sparse.

\subsubsection*{Conclusion}

The methods presented in this paper to derive stable and interpretable
prediction rules have shown promises in predicting preterm births.
The accuracy achieved is better than those reported in the literature.
As the classifiers are derived directly from the hospital database,
they can be implemented to augment the operational workflow. The prediction
rules can be used in paper-form as a check-list and a fast look-up
risk table.

\bibliographystyle{elsarticle-harv}

\end{document}